\documentclass[10pt,twocolumn,letterpaper]{article}

\usepackage[pagenumbers]{cvpr} 

\definecolor{cvprblue}{rgb}{0.21,0.49,0.74}
\usepackage[pagebackref,breaklinks,colorlinks,allcolors=cvprblue]{hyperref}
\usepackage{amsmath,amssymb,amsthm}
\usepackage{booktabs}
\usepackage[table,xcdraw]{xcolor}
\usepackage[normalem]{ulem}
\useunder{\uline}{\ul}{}
\usepackage{bm}
\usepackage{algorithm}
\usepackage{algorithmicx} 
\usepackage{algpseudocode}
\usepackage{makecell}

\title{MagicFuse: Single Image Fusion for Visual and Semantic Reinforcement}

\author{Hao Zhang$^{1,2}$ \thanks{Equal Contribution}\quad
	Yanping Zha$^{1}$$\,^{*}$ \quad
	Zizhuo Li$^{1}$\quad
	Meiqi Gong$^{1}$ \quad
	Jiayi Ma$^{1,3}$ \thanks{Corresponding author} \\
	$^1$Electronic Information School, Wuhan University, China\\
	$^2$Suzhou Institute of Wuhan University, China\\
	$^3$School of Automation, Wuhan University, China\\
	{\tt\small \{zhpersonalbox, jyma2010\}@gmail.com},\quad
	{\tt\small \{yanpingzha,  zizhuo\_li, meiqigong\}@whu.edu.cn}
}

\begin{document}
\maketitle
\begin{abstract}
This paper focuses on a highly practical scenario: how to continue benefiting from the advantages of multi-modal image fusion under harsh conditions when only visible imaging sensors are available. To achieve this goal, we propose a novel concept of single-image fusion, which extends conventional data-level fusion to the knowledge level. Specifically, we develop MagicFuse, a novel single image fusion framework capable of deriving a comprehensive cross-spectral scene representation from a single low-quality visible image. MagicFuse first introduces an intra-spectral knowledge reinforcement branch and a cross-spectral knowledge generation branch based on the diffusion models. They mine scene information obscured in the visible spectrum and learn thermal radiation distribution patterns transferred to the infrared spectrum, respectively. Building on them, we design a multi-domain knowledge fusion branch that integrates the probabilistic noise from the diffusion streams of these two branches, from which a cross-spectral scene representation can be obtained through successive sampling. Then, we impose both visual and semantic constraints to ensure that this scene representation can satisfy human observation while supporting downstream semantic decision-making. Extensive experiments show that our MagicFuse achieves visual and semantic representation performance comparable to or even better than state-of-the-art fusion methods with multi-modal inputs, despite relying solely on a single degraded visible image. The code is publicly available at~\url{https://github.com/zhayanping/MagicFuse}.
\end{abstract}

\section{Introduction}
\label{sec:intro}
Visible imaging sensors struggle to comprehensively capture scene content under harsh conditions, often yielding degraded appearances, \textit{e.g.}, low-light, hazy, or noised. In this context, the infrared and visible image fusion (IVIF) technique has emerged. It leverages the complementary information from infrared modality to compensate for visible ones, enhancing the completeness of scene representation for human observation and machine perception~\cite{zhang2023visible, zhang2021image}. Owing to its excellent characteristics, this technique has been widely applied in various human-/machine-centered perception systems, \textit{e.g.}, individual reconnaissance, intelligent transportation, and assisted driving.

\begin{figure}[!t]
	\centering
	\includegraphics[width=1\linewidth]{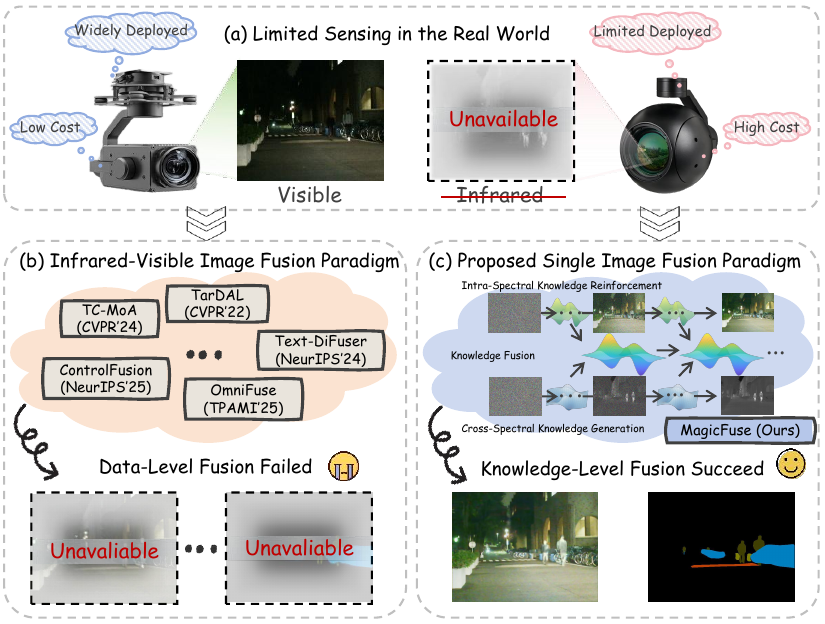}
	\caption{Under limited sensing conditions, existing image fusion methods fail completely, whereas our MagicFuse still achieves promising cross-spectral scene representation.}\label{fig:1}
\end{figure}
In recent years, deep learning has greatly advanced IVIF technology, continuously pushing its adaptability to complex real-world environments to new heights. Typically, TarDAL~\cite{liu2022target} employs a detection-driven dual adversarial learning architecture to produce salient target-aware fused images, which are particularly effective in enhancing the accuracy of semantic decision tasks under complex environments. OmniFuse~\cite{zhang2025omnifuse} employs the diffusion model to perform degradation removal and information fusion in the latent space, enabling visually clear results even under challenging environments with composite degradations. Cleverly, Text-DiFuse~\cite{zhang2024text} introduces a text-modulated robust image fusion model. Guided by user instructions, it enhances both the visual quality and semantic expressiveness of fused images under harsh imaging conditions.

Admittedly, these IVIF methods offer attractive advantages. They generally rely on a strict prerequisite: paired visible and infrared images must be simultaneously available~\cite{xu2023murf,li2025mulfs}. However, a noteworthy reality is that in most real-world scenarios, only visible imaging sensors are available, while thermal infrared sensors are absent due to their high cost, as shown in Fig.~\ref{fig:1} (a). \textit{Such a data-level absence fundamentally invalidates all existing IVIF methods}, as shown in Fig.~\ref{fig:1} (b). Although image restoration techniques~\cite{conde2024instructir, luo2023image} can be employed to recover as much scene information as possible from the visible spectrum, the prior knowledge within a single spectrum remains limited. When degradation types are diverse and coupled, existing restoration methods often fail to achieve satisfactory results.

It is natural to pose the following bold idea: \textit{could we continue to enjoy the benefits of IVIF in harsh environments with only visible imaging sensors}? This is highly attractive, as it allows for a comprehensive cross-modal scene representation from a single low-quality visible image alone. This motivates our investigation into a previously unexamined research direction: \textbf{Single Image Fusion} (SIF). Before proceeding, we must raise two fundamental questions of SIF. (1) To overcome the representational limits of visible images, where does the new knowledge beyond the visible spectrum originate?  (2) Since data-level fusion is infeasible due to the absence of infrared data, how can we achieve a leap from data-level to knowledge-level fusion?

Driven by these two fundamental questions, we propose MagicFuse, a single-image fusion framework that infers a joint visible–infrared scene representation from a single low-quality visible image, enhancing both visual quality and semantic decision accuracy, as shown in Fig.~\ref{fig:1} (c). First, powerful generative models may help us address the first question, as they are capable of producing new knowledge through learning from large-scale data. Accordingly, we design an intra-spectral knowledge reinforcement (IKR) branch and a cross-spectral knowledge generation (CKG) branch based on the diffusion model. IKR is dedicated to enhancing scene information from low-quality visible images, with the goal of maximizing knowledge mining within the visible spectrum. In contrast, CKG aims to learn thermal radiation distribution patterns from large-scale paired multi-modal images, thereby generating new knowledge in the infrared spectrum from the visible one. Second, to answer the second question, we consider that the estimated noise at each timestep in the diffusion sampling process inherently encodes valuable knowledge, which can be exploited as a medium for knowledge fusion. Therefore, we develop a multi-domain knowledge fusion (MKF) branch to integrate the probabilistic noise from diffusion streams of IKR and CKG branches. By performing continuous sampling from a shared starting point (\textit{i.e.}, standard Gaussian noise), the MKF branch finally generates a visual image with cross-spectral representation capability, called Magic Image (\textbf{MagImg}). To further enhance the compatibility of MagImg for downstream semantic decision tasks, we introduce an auxiliary segmentation head embedded into the MKF branch. It aligns the fused features with semantic labels, enriching MagImg with refined structural semantics. Therefore, our MagicFuse constructs enhanced cross-spectral scene representations from a single low-quality visible image by mining new knowledge and performing knowledge fusion, improving both visual and semantic perception of degraded scenarios.

In summary, our contributions are as follows:
\begin{itemize}
	\item We propose MagicFuse, a novel single image fusion framework capable of deriving a comprehensive cross-spectral scene representation from a single low-quality visible image. To our knowledge, this is the first proposal of SIF, offering significant potential to broaden the practical scope of image fusion.
    \item We leverage IKR and CKG for intra-spectral knowledge reinforcement and cross-spectral knowledge generation, and integrate them with MKF to bridge data-level to knowledge-level fusion, thereby addressing two fundamental questions of SIF.
    \item We construct visually and semantically coupled scene representation constraints, enabling our MagicFuse to simultaneously preserve both visual and semantic perception capabilities for degraded scenarios.
	\item Extensive experiments confirm our MagicFuse's reliability and practicality, achieving highly competitive performance in both visual and semantic perception.
\end{itemize}

\begin{figure*}[t]
	\centering
	\includegraphics[width=1\linewidth]{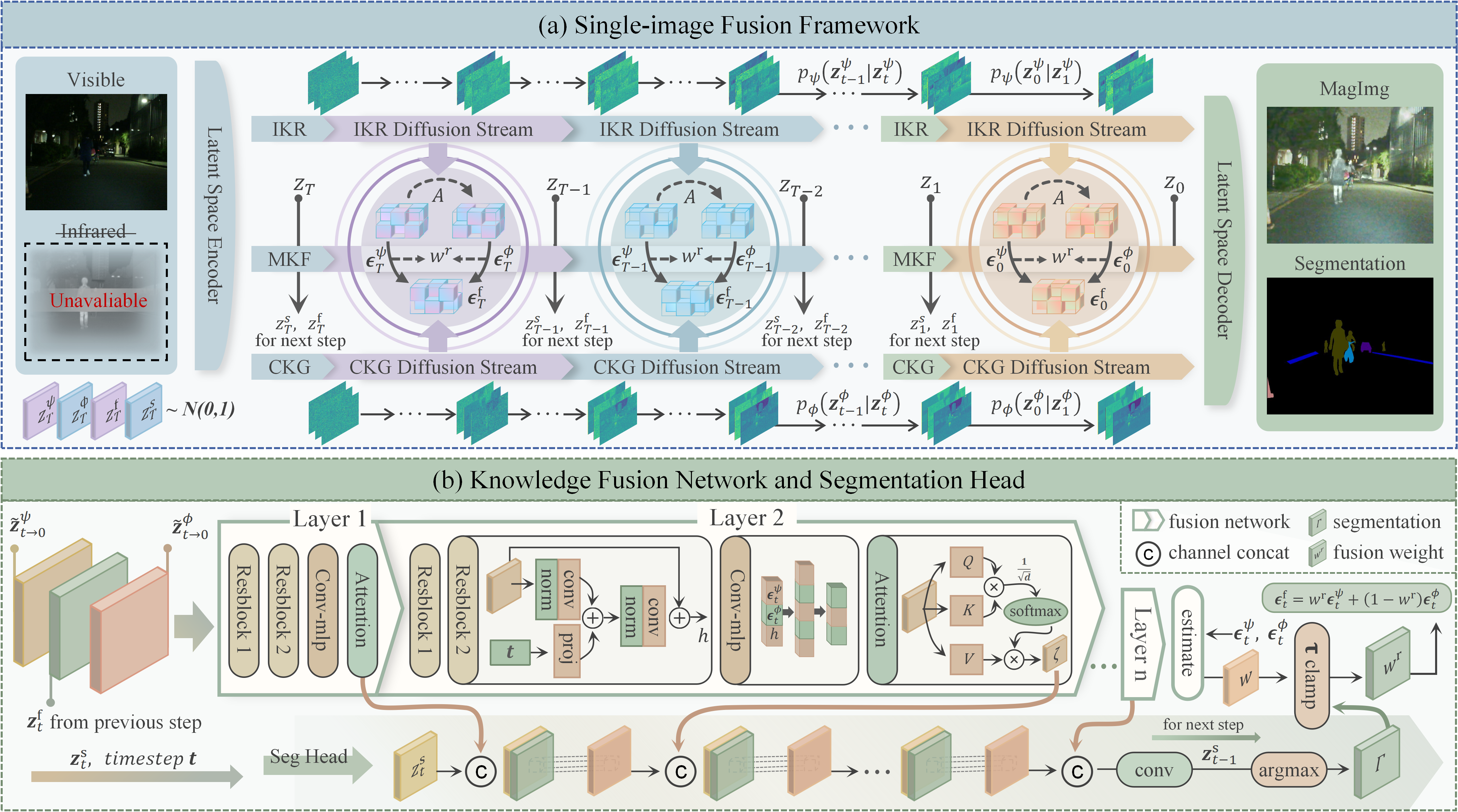}
	\caption{Overall Framework of our proposed MagicFuse.}\label{fig:2}
\end{figure*}

\section{Methodology}
\subsection{Problem Formulation}
Our MagicFuse aims to obtain a cross-spectral scene representation through knowledge fusion using only single degraded visible images. The overall framework is presented in Fig.~\ref{fig:2}. Two types of knowledge are considered. Intra-spectral knowledge captures color, texture, and other details obscured by degradations, reflecting inherent visible-spectrum characteristics. We adopt image restoration as its task carrier, leveraging a diffusion model to recover weakened or lost information, thereby reinforcing the visible-spectral knowledge. Cross-spectral knowledge learns thermal radiation patterns from the visible spectrum, inferring object-level radiation intensity and energy distribution. This is achieved via visible-to-infrared translation, where a diffusion model progressively infers latent infrared representations, generating new cross-spectral knowledge. Thus, the SIF problem can be formulated as a knowledge fusion task:
\begin{equation}
	\begin{split}
		&\;\;\min\limits_{\bm{\omega}^{\mathtt{f}}} \mathbb{E}_{\bm{z}_T \sim \mathcal{N}(0,\bm{I})} \left[\mathcal{L}^{\mathtt{f}}\big(\mathcal{F}(\bm{k}^\psi,\bm{k}^\phi;\bm{\omega}^{\mathtt{f}})\big)  \right] ,\\
		&s.t. \hspace{4pt} \bm{k}^\psi\in\Psi(\bm{\mathcal{I}},\bm{z}_T; \bm{\omega}^\psi),  \;\bm{k}^\phi\in\Phi(\bm{\mathcal{I}},\bm{z}_T; \bm{\omega}^\phi).
	\end{split} \label{eq:1}
\end{equation}
where $\mathcal{L}^{\mathtt{f}}$ denotes the fusion loss, and $\mathcal{F}(\cdot; \bm{\omega}^{\mathtt{f}})$ represents the knowledge fusion network in the MKF branch parameterized by $\bm{\omega}^{\mathtt{f}}$. $\bm{\mathcal{I}} \in \mathbb{R}^{H \times W \times 3}$ denotes the input degraded visible image. $\Psi(\cdot; \bm{\omega}^\psi)$ and $\Phi(\cdot; \bm{\omega}^\phi)$ indicate two diffusion streams for visible image restoration and visible-to-infrared image translation, with learnable parameters $\bm{\omega}^\psi$ and $\bm{\omega}^\phi$, respectively. $\bm{k}^\psi$ and $\bm{k}^\phi$ correspond to the above two types of knowledge produced by $\Psi$ and $\Phi$ at different timesteps. For a total of $T$ timesteps, $\bm{k}^\psi \in \{k^\psi_1, k^\psi_2, \dots, k^\psi_T\}$ and $\bm{k}^\phi \in \{k^\phi_1, k^\phi_2, \dots, k^\phi_T\}$. $\bm{z}_T$ is the noise sampled from a standard Gaussian distribution, serving as the shared starting point of the sampling process in $\Psi$ and $\Phi$.

By solving \cref{eq:1}, we can obtain a pleasing MagImg that enables cross-spectral visual representation of degraded scenarios. However, recent studies~\cite{tang2022image} generally agree that image fusion should not only focus on visual quality for human perception but also ensure semantic attributes for machine understanding. To achieve this goal, we embed a segmentation head in the MKF branch to enforce alignment between the fused features and semantic labels, thereby enriching the semantic information in the knowledge fusion process. Consequently, \cref{eq:1} can be revised as follows:
\begin{equation}
	\label{eq:2}
	\min_{\bm{\omega}^{\mathtt{f}}, \bm{\omega}^{\mathtt{s}}} \left[ \mathcal{L}^{\mathtt{f}}\big( \mathcal{F}(\bm{k}^\psi,\bm{k}^\phi; \bm{\omega}^{\mathtt{f}}) \big) + \mathcal{L}^{\mathtt{s}}\big( \mathcal{S}(\bm{\zeta}; \bm{\omega}^{\mathtt{s}}) \big) \right],
\end{equation}
where $\mathcal{S}(\cdot; \bm{\omega}^{\mathtt{s}})$ represents the segmentation head parameterized by $\bm{\omega}^{\mathtt{s}}$. $\bm{\zeta}$ denotes the attention features extracted from fusion network $\mathcal{F}$, which are serve as inputs to the segmentation head $\mathcal{S}$, ensuring effective coupling of visual and semantic features. Solving \cref{eq:2} enables knowledge fusion, obtaining a cross-spectral scene representation that preserves both visual and semantic performance from a single low-quality visible image. 

\subsection{Knowledge Reinforcement and Generation}
In the IKR and CKG branches, we employ the latent diffusion models (LDMs) $\Psi$ and $\Phi$ to reinforce intra-spectral knowledge and generate cross-spectral knowledge. Note that $\Psi$ and $\Phi$ share the same basic LDM design but differ in parameter capacity. This choice is motivated by the differing difficulty levels of the image restoration and visible-to-infrared image translation tasks. Below, we uniformly describe their architectures.

\noindent\textbf{Autoencoder Mapping.} 
We introduce lightweight autoencoders for $\Psi$ and $\Phi$, which incorporate InstanceNorm to guide the network in learning content-centric features, promoting a smooth and continuous manifold structure in the latent space. This resulting architecture facilitates more stable and efficient diffusion model training and better suits the requirements of image restoration and visible-to-infrared image translation tasks. Uniformly, the encoding and decoding processes are formalized as: $\bm{z}=\bm{\mathcal{E}}(\bm{\mathcal{I}}), \hat{\bm{\mathcal{I}}}=\bm{\mathcal{D}}(\bm{z})$.

\noindent\textbf{Diffusion Estimation.} 
We employ a latent diffusion model built upon a U-Net backbone. The model operates by gradually corrupting the latent representation $\bm{z}_0$ of the target image into Gaussian noise over $T$ timesteps (forward process) and then learning to reverse this process to reconstruct $\bm{z}_0$ from pure noise $\bm{z}_T \sim \mathcal{N}(\mathbf{0}, \bm{I})$ (reverse process). Instead of modeling the complex data distribution directly, the network is trained to predict the added noise $\bm{\epsilon}_{\theta}(\bm{z}_t,t)$ via a simplified variational objective~\cite{ho2020denoising}. 

To adapt this framework for image restoration and visible-to-infrared translation, we introduce task-specific conditioning: the degraded visible image serves as the condition $\bm{c}$, concatenated with the noisy latent $\bm{z}_t$ along the channel dimension. Consequently, the denoiser predicts the noise component as $\bm{\epsilon}_\theta(\bm{z}_t, \bm{c}, t)$.  During inference, we adopt the accelerated deterministic sampling strategy from DDIM~\cite{song2021denoising} to efficiently recover $\bm{z}_0$ from $\bm{z}_T$. Here, $\bm{z}_0$ denotes the latent representation of the clean visible or infrared image depending on the task.

Image restoration and visible-to-infrared translation exhibit fundamentally different learning objectives: the former focuses on recovering intra-spectral details, while the latter requires modeling cross-spectral semantic correspondence. 
To explicitly capture these complementary properties, we instantiate two specialized diffusion branches built upon the same generative framework. Specifically, the IKR and CKG branches are designed to learn \textit{intra-spectral reinforced knowledge} ($\bm{k}^\psi$) and \textit{cross-spectral generated knowledge} ($\bm{k}^\phi$), respectively. 
This is achieved by training two separate LDMs ($\Psi$ and $\Phi$) on paired degraded-clean and visible-infrared datasets. The acquired knowledge is implicitly embedded in the noise predicted during the denoising process: $\bm{\epsilon}^{\psi}_t = \bm{\epsilon}^\psi_{\theta}(\bm{z}^\psi_t, \bm{c}, t)$, $\bm{\epsilon}^{\phi}_t =\bm{\epsilon}^\phi_{\theta}(\bm{z}^\phi_t, \bm{c}, t)$. 

\begin{algorithm}[t]
	\caption{Optimization Scheme of our MagicFuse}
	\label{algorithm1}
	\begin{algorithmic}[1]
		\Require 
		Training dataset $\mathcal{R}$; 
		Diffusion steps $T$;
		Diffusion model $\Psi$ in IKR; 
		Diffusion model $\Phi$ in CKG;
		Autoencoder $\{\bm{\mathcal{E}}, \bm{\mathcal{D}}\}$;
		Knowledge fusion network $\mathcal{F}$;
		Segmentation head $\mathcal{S}$
		\Ensure 
		Optimized parameters $\bm{\omega}^\mathtt{f}, \bm{\omega}^\mathtt{s}$
		\State Initialize $\bm{\omega}^\mathtt{f}, \bm{\omega}^\mathtt{s}$ randomly
		\For{$epoch = 1$ to $N$}
		\For{$(\bm{\mathcal{I}}, \hat{\Gamma}) \in  \mathcal{R}$}  \Comment{$\hat{\Gamma}$ is segmentation label}
		\State Sample $\bm{z}_T^{\{\psi,\phi,\bm{\mathtt{f}}\}} \sim \mathcal{N}(0, \bm{I})$
		\Comment{$\bm{z}_T^{\psi}=\bm{z}_T^{\phi}=\bm{z}_T^{\bm{\mathtt{f}}}$}
		\State Sample $\bm{z}_T^{\bm{\mathtt{s}}} \sim \mathcal{N}(0, \bm{I})$ 
		\Comment{$\in \hspace{-2pt}\mathbb{R}^{B\times N_{class}\times H \times W}$}
		\State Select $\bm{i}$ from $[0, T]$ randomly
		\For{$t = T$ downto $\bm{i}$}  \Comment{Reverse process}
		\State $\bm{z}_t^{\psi}, \bm{\epsilon}_t^{\psi}, \widetilde{\bm{z}}_{t\to0}^{\psi} \gets \Psi(\bm{\mathcal{I}}, \bm{z}_t^{\psi}, t)$
		\State $\bm{z}_t^{\phi}, \bm{\epsilon}_t^{\phi}, \widetilde{\bm{z}}_{t\to0}^{\phi} \gets \Phi(\bm{\mathcal{I}}, \bm{z}_t^{\phi}, t)$
		\State  \text{Enable gradient for $\mathcal{F}(\cdot)$, $\mathcal{S}(\cdot)$ when $t = i$}
		\State $ \bm{w}^{\bm{\mathtt{r}}} \gets \mathcal{F}( \widetilde{\bm{z}}_{t\to0}^{\psi},\widetilde{\bm{z}}_{t\to0}^{\phi},\bm{z}_t^{\bm{\mathtt{f}}},\bm{\epsilon}_t^{\psi},\bm{\epsilon}_t^{\phi}, t)$
		\State $\bm{z}_t^{\bm{\mathtt{s}}}, \Gamma \gets \mathcal{S}(\bm{z}_t^{\bm{\mathtt{s}}},\zeta)$
		\Comment {$\zeta \;\text{from} \;\mathcal{F}(\cdot)$}
		\State $\bm{\epsilon}_t^{\bm{\mathtt{f}}} \gets \bm{\epsilon}_t^{\psi},\bm{\epsilon}_t^{\phi},\bm{w}^{\bm{\mathtt{r}}}$
		\State $\bm{z}_t^{\bm{\mathtt{f}}} \gets p(\bm{z}_{t-1}^{\bm{\mathtt{f}}}|\bm{z}_t^{\bm{\mathtt{f}}}, \widetilde{\bm{z}}_{t\to0}^{\mathtt{f}}, \bm{\epsilon}_t^{\bm{\mathtt{f}}} )$
		\EndFor
		\State $\{\bm{I}_{\psi},\bm{I}_{\phi},\bm{I}_{\text{MImg}}\} \gets \bm{\mathcal{D}}(\{\widetilde{\bm{z}}_{t\to0}^{\psi},\widetilde{\bm{z}}_{t\to0}^{\phi},\widetilde{\bm{z}}_{t\to0}^{\bm{\mathtt{f}}}\})$
		\State $\text{update } \bm{\omega}^\mathtt{f}, \bm{\omega}^\mathtt{s} \text{ by } \mathcal{L}^{\mathtt{f}}(\bm{I}_{\psi},\bm{I}_{\phi},\bm{I}_{\text{MImg}}), \mathcal{L}^{\mathtt{s}}(\bm{z}_t^{\bm{\mathtt{s}}}, \hat{\Gamma})$
		\EndFor
		\EndFor
	\end{algorithmic}
\end{algorithm}

\subsection{Knowledge Fusion}
To fuse knowledge $\bm{k}^\psi$ and $\bm{k}^\phi$, we construct an MKF branch aligned with the diffusion streams of the IKR and CKG branches. As shown in Fig.~\ref{fig:2} (b), a knowledge fusion network $\mathcal{F}$ is designed to generate the weighting coefficients, combining the noise estimated by denoisers in $\Psi$ and $\Phi$:
\begin{equation}
	\bm{\epsilon}_t^\mathtt{f} = \bm{w} \bm{\epsilon}^{\psi}_t + (1 - \bm{w}) \bm{\epsilon}^{\phi}_t. \label{eq:8}
\end{equation}
The rationality of the weighting coefficient is crucial, as it determines whether the knowledge $\bm{k}^\psi$ and $\bm{k}^\phi$ are balanced in the probabilistic space. We consider the generation of $\bm{w}$ from two perspectives. First, the knowledge quality of the diffusion streams in the IKR and CKG branches needs to be evaluated, which can be achieved by comparing their respective estimated representations $\bm{\widetilde{z}}^{\psi}_{t\to0}$ and $\bm{\widetilde{z}}^{\phi}_{t\to0}$. Second, multi-domain noise $\bm{\epsilon}^{\psi}_t$ and $\bm{\epsilon}^{\phi}_t$ should be analyzed, since they are the objects of weight aggregation. Besides, to satisfy the iterative nature of the diffusion stream, the current-step sampled representation $\bm{z}_t^\mathtt{f}$ in the MKF branch also needs to participate in the weighting estimation. Therefore, the computation of $\bm{w}$ in \cref{eq:8} can be formalized as:
\begin{equation}
	\bm{w} = \mathcal{F}(\bm{\widetilde{z}}^{\psi}_{t\to0}, \bm{\widetilde{z}}^{\phi}_{t\to0}, \bm{{z}}_t^\mathtt{f}, \bm{\epsilon}^{\psi}_t, \bm{\epsilon}^{\phi}_t, t), \label{eq:9}
\end{equation}
where both $\bm{\widetilde{z}}^{\psi}_{t\to0}$ and $\bm{\widetilde{z}}^{\phi}_{t\to0}$ are calculated according to $\bm{\widetilde{z}}_{t\to0} = \frac{\bm{z}_t - \sqrt{1 - \bar{\alpha}_t} \, \bm{\epsilon}_t}{\sqrt{\bar{\alpha}_t}}$. Notably, $\bm{\widetilde{z}}_{t\to0}$ is not the result of full sampling from the diffusion model, but rather a single-step deterministic estimate of the initial representation.

\begin{figure*}[t]
	\centering
	\includegraphics[width=1\linewidth]{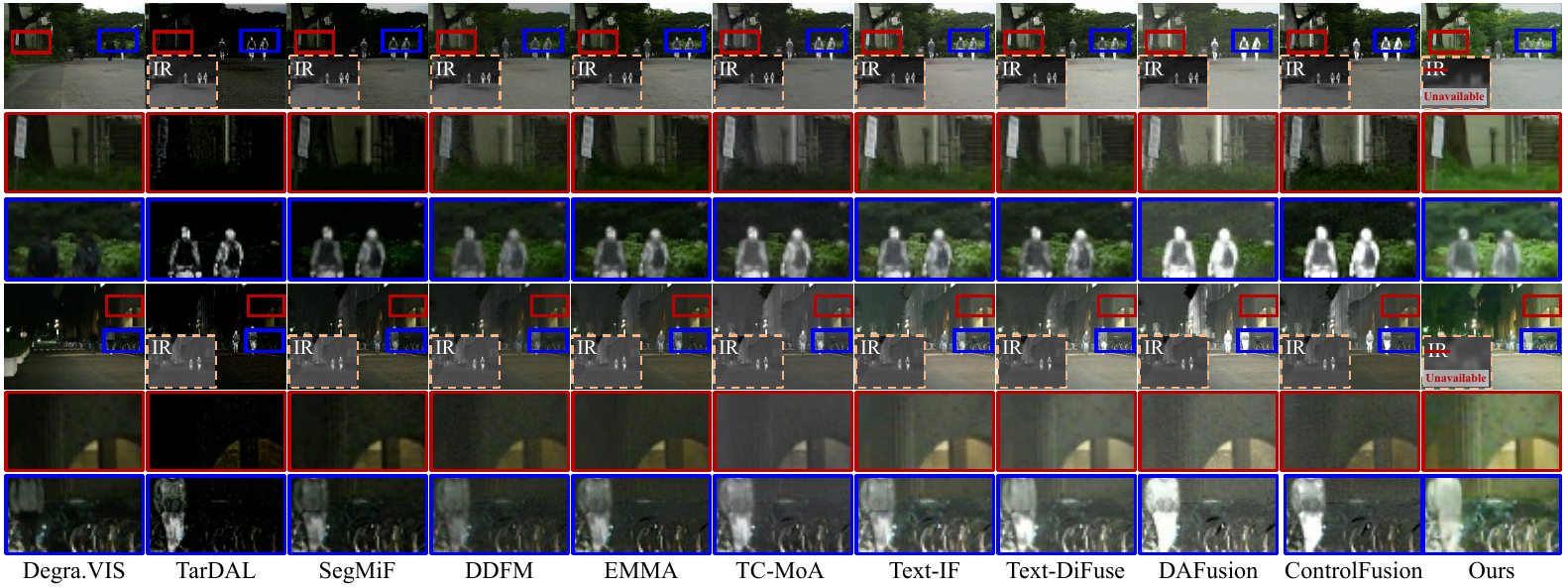}
	\vspace{-0.25in}
	\caption{Qualitative visual representation comparison.}\label{fig:3}
	\vspace{-0.05in}
\end{figure*}
\begin{table*}[t]
	\caption{Quantitative visual comparison. All competitors use infrared-visible pairs, while ours uses only single visible images.} \label{tab:2}
	\vspace{-0.1in}
	\centering
	\scriptsize 
	\setlength{\tabcolsep}{9.2pt}
	\renewcommand{\arraystretch}{0.9}
	\begin{tabular}{lcccccccccc}
		\toprule
		\textbf{Visual} & \textbf{TarDAL} & \textbf{SegMiF} & \textbf{DDFM} & \textbf{EMMA} & \textbf{TC-MoA} & \textbf{Text-IF} & \textbf{Text-DiFuse} & \textbf{DAFusion} & \textbf{ControlFusion} & \textbf{Ours} \\ 
		\midrule
		\textbf{EN} $\uparrow$ & 5.11 & 6.40 & 6.59 & 6.74 & 6.63 &6.96 & 7.08 & \cellcolor[HTML]{E6E6FF}{\ul 7.26} & 7.00 & \cellcolor[HTML]{FFE6E6}\textbf{7.29} \\
		\textbf{SSIM}  $\uparrow$ & 0.12 & 0.29 & 0.36 & 0.43 & \cellcolor[HTML]{E6E6FF}{\ul 0.49} & \cellcolor[HTML]{FFE6E6}\textbf{0.52} & 0.41 & 0.42 & 0.43 & 0.45 \\
		\textbf{MI}  $\uparrow$ & 1.45 & 2.69 & 2.68 & \cellcolor[HTML]{E6E6FF}{\ul 3.46} & 2.94 & 3.31 & 2.99 & 3.08 & 3.38 & \cellcolor[HTML]{FFE6E6}\textbf{4.13} \\
		\textbf{Qabf}  $\uparrow$ & 0.21 & 0.38 & 0.31 & 0.45 & 0.47 & \cellcolor[HTML]{FFE6E6}\textbf{0.59} & 0.40 & 0.33 & 0.45 & \cellcolor[HTML]{E6E6FF}{\ul 0.50} \\
		\textbf{PSNR}  $\uparrow$ & 57.48 & 59.45 & 62.15 & 61.96 & 63.03 & \cellcolor[HTML]{E6E6FF}{\ul 63.38} & 62.99 & 61.71 & 61.68 & \cellcolor[HTML]{FFE6E6}\textbf{63.49} \\
		\bottomrule
	\end{tabular}
\end{table*}
\subsection{Visual-Semantic Reinforcement}
\textbf{Visual-Semantic Coupling.}   
We expect the guidance of knowledge fusion to encompass not only the visual aspect but also the semantic one. Therefore, a segmentation head $\mathcal{S}$ is embedded in the MKF branch. As illustrated in Fig.~\ref{fig:2} (b), in addition to the implicit representation $\bm{z}^{s}_t$ from the previous step for segmentation, $\mathcal{S}$ primarily takes the attention features $\bm{\zeta}$ from the knowledge fusion network as input:
\begin{equation}
	\Gamma = \arg\max(\;\underbrace{\mathcal{S}(\bm{z}^{s}_t, \zeta_1, \zeta_2, \dots, \zeta_l)}_{\bm{z}^{s}_{t-1} \text{ for next step}}\;),\label{eq:10}
\end{equation}
where $\Gamma$ is the predicted segmentation map, and $\zeta_l$ denotes the attention map from the $l$-th layer of $\mathcal{F}$. Such an operation is designed to enhance the semantic quality of visual features under the guidance of segmentation optimization. Its rationality can be evidenced from~\cite{kim2025seg4diff}, as sufficiently general visual features are capable of supporting semantic decisions. In turn, a radiation category map $\mathcal{M}$ derived from $\Gamma$ is used to further refine the weighting coefficients:
\begin{equation}
	\bm{w}^{\mathtt{r}}=\mathcal{M}\,\text{min}(\bm{w}, \bm{\tau})+(1-\mathcal{M})\,\bm{w}, \label{eq:11}
\end{equation}
where $\mathcal{M}$ indicates typical thermal categories (\textit{e.g.}, pedestrians, cars), used to enhance the preservation of thermal objects during the fusion process. $\bm{\tau}$ serves as a hyperparameter to regulate the impact of the radiation category map. 

Notably, the above enforced modulation not only incorporates segmentation into the knowledge fusion process to foster visual–semantic coupling but also resolves a key optimization challenge in knowledge fusion. More concretely, since the estimated noise $\bm{\epsilon}^{\psi}_t$ and $\bm{\epsilon}^{\phi}_t$ are derived from the same input $\{\bm{\mathcal{I}}, \bm{z}_T\}$, an inherent correlation exists between them: $\bm{\epsilon}^{\phi}=A\,\bm{\epsilon}^{\psi}$, where $A$ is a transformation matrix.  \cref{eq:8} can be reformulated as $\bm{\epsilon}_t^\mathtt{f} = (\bm{w}+(1-\bm{w})A) \bm{\epsilon}^{\psi}_t$. At this time, given that the input conditions in both $\Psi$ and $\Phi$ are the visible images, the optimization tends to be completely biased towards $\bm{\epsilon}^{\psi}_t$, which is manifested in $\bm{w}\to1$, leading to training collapse. The enforced modulation based on $\mathcal{M}$ in \cref{eq:11} effectively mitigates this problem, facilitating rational knowledge fusion. Therefore, \cref{eq:8} can be rewrite as: $\bm{\epsilon}_t^\mathtt{f} = \bm{w}^{\mathtt{r}} \bm{\epsilon}^{\psi}_t + (1 - \bm{w}^{\mathtt{r}}) \bm{\epsilon}^{\phi}_t$.

\noindent\textbf{Visual and Semantic Derivation.} 
In the MKF branch, for any $t\hspace{-2pt}\in \hspace{-2pt} \{T, \cdots, 1\}$, we first perform continuous iterative sampling from timestep $T$ to $t$:
\begin{equation}
	\begin{gathered}
		\bm{z}_{T-1}^{\mathtt{f}} = \sqrt{\bar{\alpha}_{T-1}}\bm{\widetilde{z}}_{T\to0}^{\mathtt{f}} + \sqrt{1-\bar{\alpha}_{T-1}}\bm{\epsilon}_T^\mathtt{f}\\
		\cdots \\
		\bm{z}_{t}^{\mathtt{f}} = \sqrt{\bar{\alpha}_{t}}\bm{\widetilde{z}}_{t+1\to0}^{\mathtt{f}} + \sqrt{1-\bar{\alpha}_{t}}\bm{\epsilon}_{t+1}^\mathtt{f}.
	\end{gathered} \label{eq:12}
\end{equation}
Then, we can perform a single-step deterministic estimate of the initial representation from $\bm{z}_{t}^{\mathtt{f}}$:
\begin{equation}
	\bm{\widetilde{z}}_{t\to0}^{\mathtt{f}} = \frac{\bm{z}_t^{\mathtt{f}} - \sqrt{1 - \bar{\alpha}_t} \, \bm{\epsilon}_{t}^\mathtt{f}}{\sqrt{\bar{\alpha}_t}}. \label{eq:13}
\end{equation}
Finally, the fused representation is mapped back to the image space via the decoder to yield the   MagImg:
\begin{equation}
	I_\text{MImg} = \bm{\mathcal{D}}(\bm{\widetilde{z}}_{t\to0}^{\mathtt{f}} ). \label{eq:14}
\end{equation}
High-quality cross-spectral visual representation of $I_\text{MImg}$ is ensured by considering its contrast, texture, and color:
\begin{equation}
	\label{eq:15}
	\left\{
	\begin{aligned}
		&\mathcal{L}_\text{cont} = \|I_\text{MImg}^y - \max(\mathcal{D}(\bm{\widetilde{z}}^{\psi}_{t\to0})^y,\mathcal{D}(\bm{\widetilde{z}}^{\phi}_{t\to0})^y)\|,  \\
		&\mathcal{L}_\text{text} \!=\!\!\| \nabla\! I_\text{MImg}^y \!\!-\!\! \max(\nabla\! \mathcal{D}(\bm{\widetilde{z}}^{\psi}_{t\to0})^y,\! \nabla\! \mathcal{D}(\bm{\widetilde{z}}^{\phi}_{t\to0})^y)\|, \\
		&\mathcal{L}_\text{color} = \|I_\text{MImg}^{cbcr} - \mathcal{D}(\bm{\widetilde{z}}^{\psi}_{t\to0})^{cbcr} \|, 
	\end{aligned}
	\right.
\end{equation}
where $y$ and $cbcr$ denote the illuminance and chrominance components, and $\nabla$ is the Sobel gradient operator. Therefore, the total visual regularization is summarized as $\mathcal{L}_\text{visual} = \lambda_1 \mathcal{L}_\text{cont} + \lambda_2\mathcal{L}_\text{text} + \lambda_3 \mathcal{L}_\text{color}$,  $\lambda_{1\sim 3}$ are used to adjust the contribution weights of these loss components.

According to \cref{eq:10}, the output $\Gamma$ of the segmentation head can be obtained. We constrain it using a cross-entropy loss to ensure the rationality of captured semantic attributes:
\begin{equation}
	\mathcal{L}_\text{seg} = -\sum_{r} \sum_{p} \hat{\Gamma}_p(r) \log \big( \Gamma_p(r) \big), \label{Eq16}
\end{equation}
where $r$ and $p$ iterate over all semantic categories and pixel positions, respectively; $\hat{\Gamma}_p(r) \in \{0,1\}$ denotes the ground truth label, and $\Gamma_p(r)$ represents the predicted probability for pixel $p$ belonging to class $r$. Finally, the joint optimization of $\mathcal{L}_\text{visual}$ and $\mathcal{L}_\text{seg}$ drives a dual enhancement of both visual fidelity and semantic consistency.

\subsection{Optimization Scheme}
We adopt a two-phase optimization strategy. At the first phase, the latent diffusion models $\Psi$ and $\Phi$ in the IKR and CKG branches are optimized independently for intra-spectral knowledge reinforcement and cross-spectral knowledge generation. At the second phase, we freeze the parameters of $\Psi$ and $\Phi$, and proceed to train the knowledge fusion network $\mathcal{F}$ along with the segmentation head $\mathcal{S}$. The training signals for this phase are derived from the outputs of the dual diffusion streams of $\Psi$ and $\Phi$ at each timestep, as outlined in \cref{algorithm1}.
\begin{figure}[!t]
	\centering
	\includegraphics[width=1\linewidth]{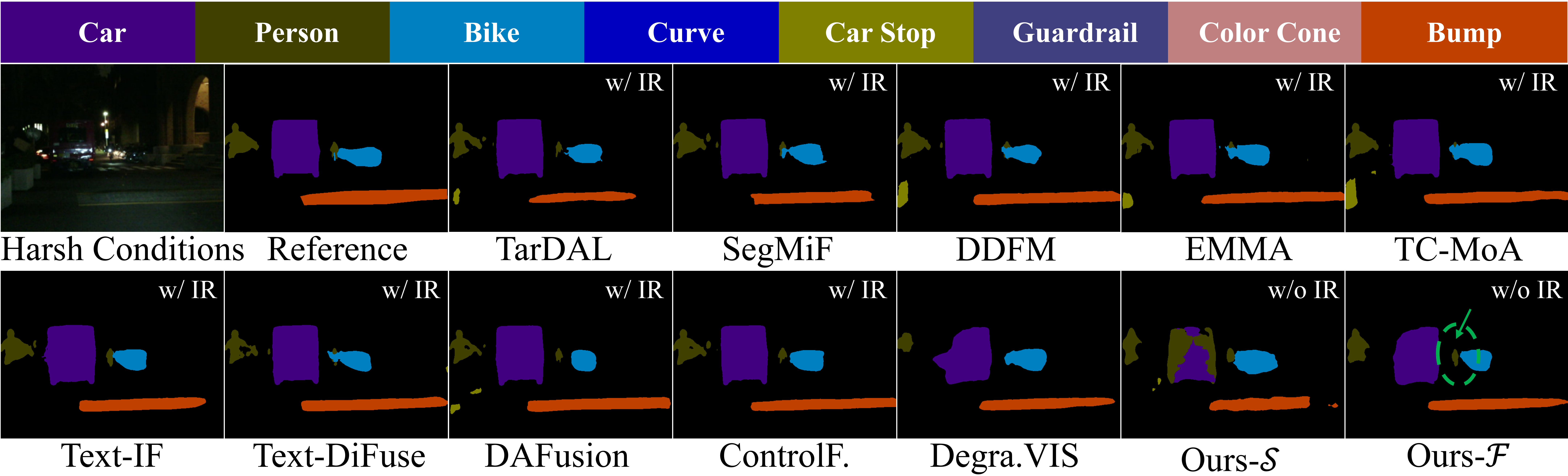}
	\vspace{-0.26in}
	\caption{Qualitative semantic representation comparison.}\label{fig:4}
\end{figure}
\begin{figure}[t]
	\centering
	\includegraphics[width=1\linewidth]{Figures/6.png}
	\vspace{-0.26in}
	\caption{Qualitative semantic representation generalization.}\label{fig:6}
	\vspace{-0.1in}
\end{figure}
\section{Experiments}
\subsection{Configuration}
\textbf{Datasets}. ${\Psi}$ and ${\Phi}$ in the IKR and CKG branches are trained on the combined set composed of MFNet~\cite{ha2017mfnet}, FMB~\cite{liu2023multi}, and LLVIP~\cite{jia2021llvip}. Specifically, we use $14,190$ degraded-clean visible image pairs to train $\Psi$ for intra-spectral knowledge reinforcement, while $25,186$ degraded visible-clean infrared image pairs are used to train $\Phi$ for cross-spectral knowledge generation. Subsequently, we use $1,177$ degraded visible images and their corresponding segmentation labels from the MFNet~\cite{ha2017mfnet} dataset for training $\mathcal{F}$ and $\mathcal{S}$ in the MKF branch. The final testing is performed on $392$ degraded visible images from the MFNet~\cite{ha2017mfnet} dataset.

\noindent{\textbf{Implementation Details.}}  ${\Psi}$, ${\Phi}$, $\mathcal{F}$, and $\mathcal{S}$ are all trained on four NVIDIA GeForce RTX 3090 GPUs with 24GB memory, and the AMD EPYC 7H12 64-core processor CPU. The training details are summarized in Table~\ref{tab:1}.

\begin{table}[t]
	\centering
	\vspace{-0.05in}
	\caption{Implementation Details of MagicFuse. Steps(T/I) denotes the number of time steps for training/inference.}\label{tab:1}
	\vspace{-0.1in}
	\centering
	\small
	\setlength{\tabcolsep}{2.9pt} 
	\renewcommand{\arraystretch}{0.9}
	\begin{tabular}{cccccccc}
	\toprule
	\textbf{\footnotesize Module}  & \textbf{\footnotesize Params(M)} & \textbf{\footnotesize FLOPs(G)}  & \textbf{\footnotesize Optimizer} & \textbf{\footnotesize LR}   & \textbf{\footnotesize BS} & \textbf{\footnotesize Steps(T/I)}   \\  \midrule
	$\bm{\Psi}$                                                                                                                                          & $39.65$ & 50.74   & Adam  & 1e-4 & $16$  & $1000/25$    \\
	$\bm{\Phi}$                                                                                                                                          & $517.67$ & 773.7  & Adam  & 1e-5 & $4$    & $1000/25$   \\
	$\bm{\mathcal{F}}\&\bm{\mathcal{S}}$    & $2.74$ &  83.60    & SGD   & 1e-3 & $1$    & $25/25$   
	\\ \bottomrule
	\end{tabular}

\end{table}

\subsection{Visual Representation Comparison}
We first evaluate the effectiveness of knowledge fusion for cross-spectral visual representation. Nine state-of-the-art fusion methods are selected to compare with our MagicFuse, including TarDAL~\cite{liu2022target}, SegMiF~\cite{liu2023multi}, DDFM~\cite{zhao2023ddfm}, EMMA~\cite{zhao2024equivariant},  TC-MoA~\cite{zhu2024task}, Text-IF~\cite{yi2024text}, Text-DiFuse~\cite{zhang2024text}, DAFusion~\cite{wang2025degradation}, and ControlFusion~\cite{tang2025controlfusion}. Notably, our MagicFuse produces the MagImg from a \textbf{\textit{single degraded visible image}}, whereas the competitors rely on \textbf{\textit{multi-modal inputs}}. Fig.~\ref{fig:3} shows that our MagicFuse effectively integrates intra-spectral reinforced knowledge and cross-spectral generated knowledge. This is reflected in its ability not only to enhance the inherent scene information contained in degraded visible images, but also to reasonably highlight important objects in the scene based on the learned thermal radiation distribution. Table~\ref{tab:2} further confirms its highly competitive performance, suggesting that \textit{even with only a single degraded visible image, it can match or surpass multi-modal image fusion methods}.

\begin{table}[t]
	\caption{Quantitative semantic comparison. All competitors use infrared-visible pairs, while ours uses only single visible images.} \label{tab:3}
	\vspace{-0.1in}
		\centering
		\scriptsize 
		\setlength{\tabcolsep}{2.2pt}
		\renewcommand{\arraystretch}{1}
		\begin{tabular}{@{}lcccccccc|c}
			\toprule
			\textbf{Semantic} & \textbf{Car} & \textbf{Person} & \textbf{Bike} & \textbf{Curve} & \textbf{Stop} & \textbf{Guar.} & \textbf{Cone} & \textbf{Bump} & \textbf{mIoU $\uparrow$}   \\ \midrule
			\textbf{TarDAL} & 85.19 & 69.39 & 70.49 & 15.89 & 45.15 & 34.83 & 51.07 & 32.13 & 55.73 \\
			\rowcolor[HTML]{FFE6E6} 
			\textbf{SegMiF} & 86.75 & 67.61 & 72.41 & 25.95 & 46.88 & 52.13 & 56.79 & 54.32 & \textbf{62.28} \\
			\textbf{DDFM} & 85.80 & 69.33 & 67.93 & 27.40 & 42.63 & 17.55 & 59.12 & 41.09 & 56.49 \\
			\textbf{EMMA} & 86.03 & 70.42 & 71.41 & 24.73 & 51.17 & 50.05 & 56.60 & 49.96 & 61.98 \\
			\textbf{TC-MoA} & 86.99 & 69.88 & 72.43 & 27.22 & 41.31 & 35.29 & 52.49 & 46.56 & 58.87 \\
			\textbf{Text-IF} & 86.25 & 69.01 & 72.05 & 28.03 & 46.76 & 43.29 & 58.79 & 44.00 & 60.65 \\
			\textbf{Text-DiFu.} & 83.24 & 68.84 & 70.93 & 25.06 & 44.13 & 43.25 & 56.56 & 49.91 & 59.93 \\
			\textbf{DAFusion} & 85.94 & 70.41 & 72.48 & 24.23 & 40.85 & 50.46 & 56.82 & 37.08 & 59.54 \\
			\textbf{ControlF.} & 86.99 & 70.50 & 71.95 & 25.66 & 42.69 & 29.32 & 53.77 & 57.39 & 59.55 \\ \midrule
			\textbf{Degra. VIS} & 81.55 & 56.35 & 70.89 & 30.82 & 40.03 & 3.89 & 61.47 & 44.51 & 54.09 \\
			\textbf{Ours-$\mathcal{S}$} & 75.93 & 59.25 & 55.83 & 41.24 & 45.77 & 59.60 & 44.34 & 35.33 & 57.11 \\
			\rowcolor[HTML]{E6E6FF} 
			\textbf{Ours-$\mathcal{F}$} & 83.80 & 61.24 & 70.62 & 35.28 & 54.09 & 49.42 & 62.69 & 45.09 & {\ul 62.19} \\ \bottomrule
		\end{tabular}
	\vspace{-0.1in}
\end{table}
\begin{figure*}[t]
	\centering
	\includegraphics[width=1\linewidth]{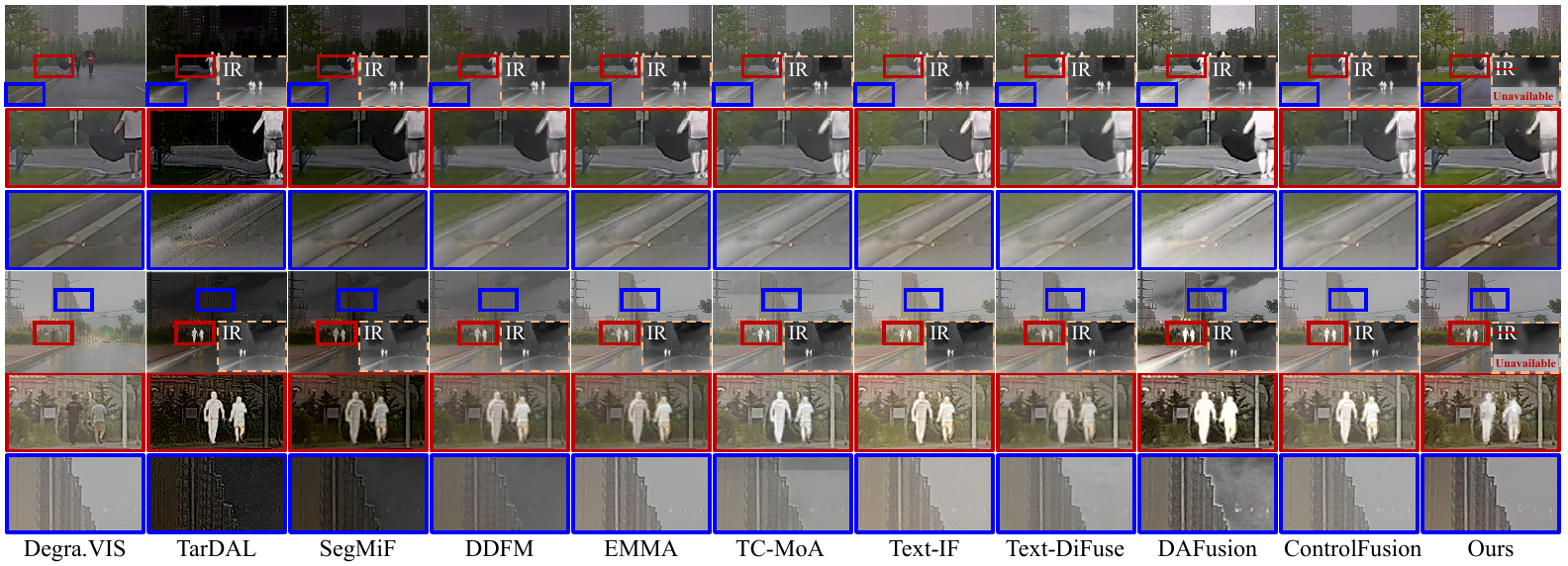}
	\vspace{-0.25in}
	\caption{Qualitative visual representation generalization.}\label{fig:5}
	\vspace{-0.05in}
\end{figure*}

\begin{table*}[t]
	\caption{Quantitative visual generalization. All competitors use infrared-visible pairs, while ours uses only single visible images.} \label{tab:4}
	\vspace{-0.1in}
	\centering
	\scriptsize 
	\setlength{\tabcolsep}{9pt}
	\renewcommand{\arraystretch}{0.9}
	\begin{tabular}{lcccccccccc}
		\toprule
		\textbf{Visual} & \textbf{TarDAL} & \textbf{SegMiF} & \textbf{DDFM} & \textbf{EMMA} & \textbf{TC-MoA} & \textbf{Text-IF} & \textbf{Text-DiFuse} & \textbf{DAFusion} & \textbf{ControlFusion} & \textbf{Ours} \\ \midrule
		\textbf{EN}$\uparrow$ & 6.44 & 6.82 & 6.71 & 6.80 & 6.76 & 6.68 & 6.93 & \cellcolor[HTML]{FFE6E6}\textbf{7.47} & 6.72 & \cellcolor[HTML]{E6E6FF}{\ul 6.93} \\
		\textbf{SSIM}$\uparrow$ & 0.19 & 0.39 & 0.46 & \cellcolor[HTML]{E6E6FF}{\ul 0.48} & 0.46 & 0.44 & 0.35 & 0.45 & \cellcolor[HTML]{FFE6E6}\textbf{0.48} & 0.40 \\
		\textbf{MI}$\uparrow$ & 2.70 & 3.59 & 3.68 & 3.99 & 3.07 & 3.71 & 3.25 & 4.05 & \cellcolor[HTML]{FFE6E6}\textbf{4.23} & \cellcolor[HTML]{E6E6FF}{\ul 4.09} \\
		\textbf{Qabf}$\uparrow$ & 0.34 & 0.60 & 0.41 & 0.66 & 0.68 & \cellcolor[HTML]{E6E6FF}{\ul 0.68} & 0.47 & 0.56 & \cellcolor[HTML]{FFE6E6}\textbf{0.70} & 0.57 \\
		\textbf{PSNR}$\uparrow$ & 60.10 & 61.99 & \cellcolor[HTML]{FFE6E6}\textbf{65.37} & 62.25 & 62.13 & 60.37 & 61.12 & 60.80 & 61.47 & \cellcolor[HTML]{E6E6FF}{\ul 62.51} \\ \bottomrule
	\end{tabular}
	\vspace{-0.1in}
\end{table*}

\subsection{Semantic Representation Comparison}
Next, we evaluate the semantic representation capability of our method. We retrain SegFormer~\cite{xie2021segformer} separately on low-quality visible images, IKR-enhanced images, fused images from comparative methods, and MagImgs from MagicFuse, to measure their contained semantic quality. Besides, we also include the auxiliary segmentation head's output for comparison. As shown in Fig.~\ref{fig:4}, under harsh conditions, SegFormer fails to detect distant pedestrians in degraded visible images. In contrast, most fusion methods enable SegFormer to detect previously missed pedestrians by leveraging complementary infrared information. Remarkably, MagicFuse achieves the same without any infrared input, using its knowledge-level understanding that pedestrians exhibit strong thermal signatures, allowing effective emphasis in the MagImg. Table~\ref{tab:3} shows that our method’s average segmentation score surpasses most multi-modal fusion methods, second only to SegMiF.

\subsection{Generalization of Knowledge Fusion}
During the learning of knowledge fusion, MagicFuse is exposed only to the MFNet~\cite{ha2017mfnet}. To verify the generalization of knowledge fusion, we directly apply the trained MagicFuse to the FMB dataset~\cite{liu2023multi}, containing $280$ test images, to assess its visual and semantic representation performance.

\begin{table}[t]
	\caption{Quantitative semantic generalization. All competitors use infrared-visible pairs, while ours uses only single visible images.} \label{tab:5}
	\vspace{-0.1in}
	\scriptsize 
	\centering
	\setlength{\tabcolsep}{1.5pt}
	\renewcommand{\arraystretch}{1}
	\begin{tabular}{lcccccccc|c}
		\toprule
		\textbf{Segmentation}            & \textbf{Car} & \textbf{Person} & \textbf{Truck} & \textbf{Lamp} & \textbf{Sign} & \textbf{Buil.} & \textbf{Vege.} & \textbf{Road} & \textbf{mIoU}$\uparrow$  \\ \midrule
		\textbf{TarDAL}               &    77.43     &      66.22      &     31.01      &     39.76     &     71.17     &     80.31      &     84.64      &     86.54     &     53.94      \\
		\rowcolor[HTML]{E6E6FF} 
		\textbf{SegMiF} &    81.94     &      70.08      &     49.89      &     44.52     &     72.20     &     82.64      &     85.89      &     88.21     &  {\ul 57.58}   \\
		\textbf{DDFM}                &    78.60     &      65.81      &     43.18      &     36.72     &     73.53     &     82.54      &     85.73      &     88.07     &     56.57      \\
		\textbf{EMMA}                &    80.71     &      68.31      &     35.29      &     42.66     &     72.84     &     81.29      &     84.45      &     86.53     &     56.89      \\
		\rowcolor[HTML]{FFE6E6} 
		\textbf{TC-MoA} &    79.92     &      66.78      &     49.28      &     41.83     &     71.08     &     81.13      &     85.70      &     88.44     & \textbf{57.72} \\
		\textbf{Text-IF}              &    81.66     &      66.73      &     22.63      &     44.14     &     71.20     &     80.37      &     85.23      &     88.87     &     56.10      \\
		\textbf{Text-DiFu.}             &    81.23     &      67.29      &     14.45      &     42.36     &     73.60     &     81.94      &     84.91      &     87.87     &     56.48      \\
		\textbf{DAFusion}              &    80.96     &      66.72      &     22.11      &     37.38     &     72.95     &     80.66      &     85.88      &     88.67     &     53.61      \\
		\textbf{ControlF.}             &    81.49     &      68.90      &     33.27      &     45.49     &     74.63     &     81.95      &     85.49      &     88.77     &     56.62      \\ \midrule
		\textbf{Degra. VIS}             &    80.72     &      56.25      &     25.89      &     28.80     &     71.11     &     79.00      &     84.58      &     89.00     &     53.39      \\
		\textbf{Ours-En. VIS}            &    80.73     &      58.41      &     33.68      &     36.41     &     73.32     &     79.21      &      84.4      &     86.39     &     54.42      \\
		\textbf{Ours-$\mathcal{F}$}         &    81.63     &      58.84      &     34.61      &     36.71     &     72.50     &     80.03      &     84.67      &     87.99     &     55.41      \\ \bottomrule
	\end{tabular}
	\vspace{-0.25in}
\end{table}

\noindent\textbf{Visual Representation Generalization}.
Fig.~\ref{fig:5} shows qualitative results of MagicFuse’s visual generalization. As can be observed, MagicFuse’s output clearly exhibits both the intra-spectral reinforced and the cross-spectral generated knowledge. Compared to degraded visible images, MagicFuse enhances scene structures, such as lane markings and building textures, while highlighting key objects like pedestrians via learned thermal patterns. Besides, the quantitative results in Table~\ref{tab:4} show that MagicFuse achieves the second-best scores on most metrics, indicating that knowledge fusion enables our method to transfer easily to unseen data, providing reliable visual representations.

\noindent\textbf{Semantic Representation Generalization}.
Fig.~\ref{fig:6} shows the qualitative results for semantic representation generalization. The segmentation map derived from the MagImg generated by our MagicFuse is highly consistent with the reference. We also provide the quantitative results in Table~\ref{tab:5}. Although our method does not surpass fusion methods with multi-modal inputs, it still significantly outperforms both the original degraded and the enhanced visible images. These results indicate that semantic attributes propagated through the knowledge fusion process remain effective on unseen data, highlighting MagicFuse’s potential for enhancing machine perception in extreme environments.

 \begin{figure}[t]
 	\centering
 	\includegraphics[width=1\linewidth]{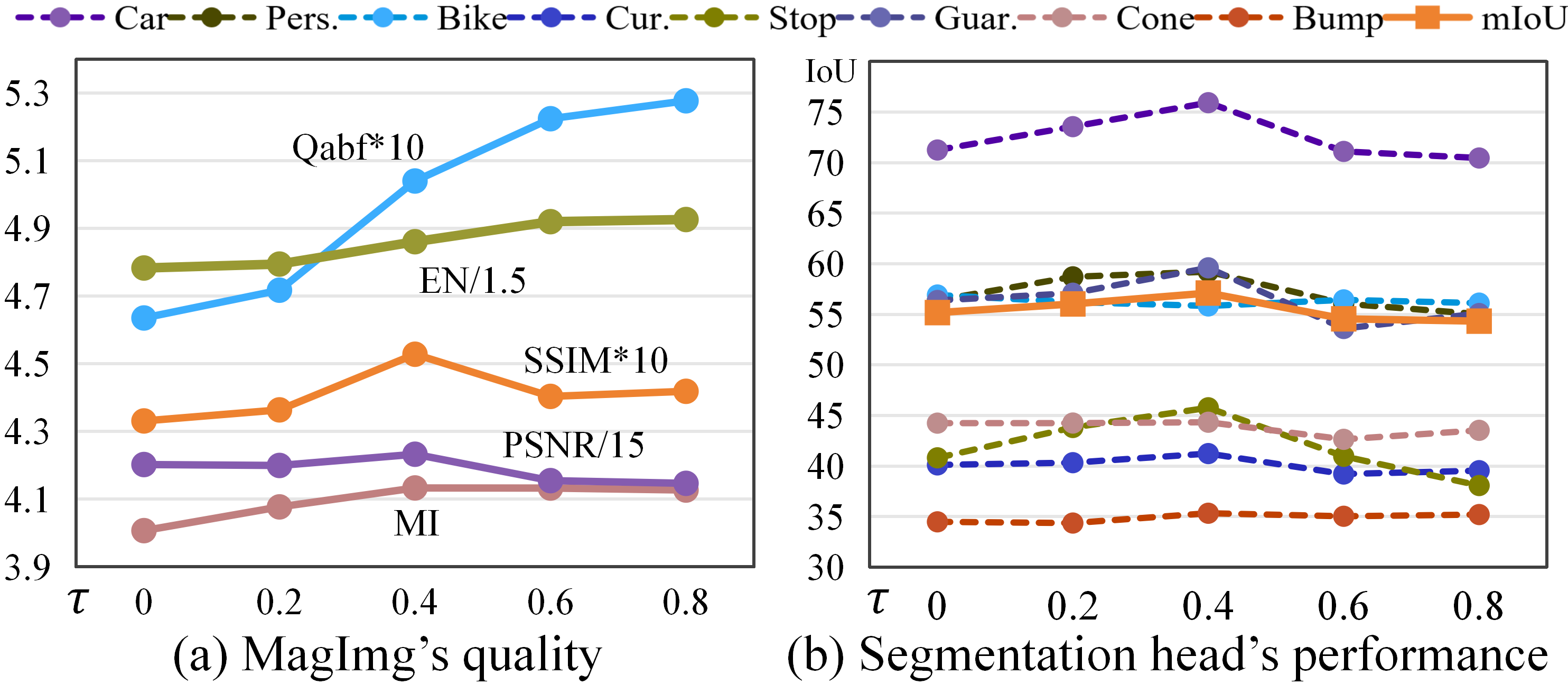}
 	\vspace{-0.26in}
 	\caption{Quantitative effects of hyperparameter $\bm{\tau}$.}\label{fig:7}
 	\vspace{-0.1in}
 \end{figure}
 
\subsection{Ablation Studies}
\textbf{Effects of Hyperparameter $\bm{\tau}$}.
The hyperparameter $\bm{\tau}$ is to regulate the influence of the radiation category map on knowledge fusion, which is also closely related to the degree of semantic attribute injection. We set $\bm{\tau}$ to $0$, $0.2$, $0.4$, $0.6$, and $0.8$, respectively, evaluating both changes of the MagImg's quality and the segmentation head's performance. The metric curves are presented in Fig.~\ref{fig:7}, where both MagImg's quality and segmentation head’s accuracy reach their highest scores when $\bm{\tau}=0.4$. This indicates that while cross-spectral generated knowledge enhances scene representation, its influence should be regulated to prevent disruption of the original spectral information.

 \begin{figure}[t]
	\centering
	\includegraphics[width=1\linewidth]{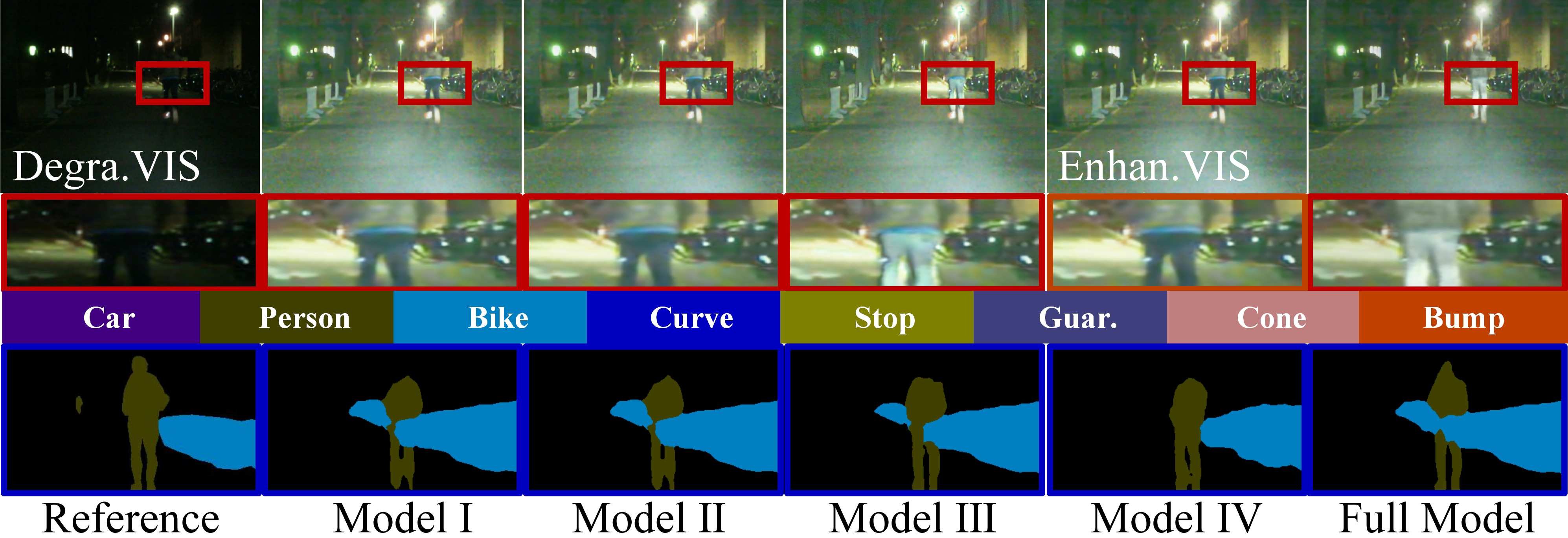}
	\vspace{-0.25in}
	\caption{Quantitative results of ablations on key components.}\label{fig:8}
	\vspace{-0.1in}
\end{figure}

\noindent\textbf{Ablations on Key Components}. 
Four model variants are derived for the ablation studies. \textit{Model I}: remove $\bm{\widetilde{z}}^{\psi}_{t\to0}$ and $\bm{\widetilde{z}}^{\phi}_{t\to0}$ from \cref{eq:9} when estimating the fusion weight. \textit{Model II}: remove the segmentation head, relying solely on visual guidance for knowledge fusion. \textit{Model III}: aggregating the outputs of the IKR and CKG branches after the diffusion process instead of fusing noise at each timestep. \textit{Model IV}: achieve scene representation based on visible images enhanced by the IKR branch. Qualitative results in Fig.~\ref{fig:8} show that Models I, II, and IV lack cross-spectral representation capability, while Model III provides only limited capability. With all components enabled, MagicFuse produces the most accurate pedestrian contours. Quantitative results in Tables~\ref{tab:6} and \ref{tab:7} confirm that removing any component degrades both visual and semantic performance.

\begin{figure}[t]
	\centering
	\includegraphics[width=1\linewidth]{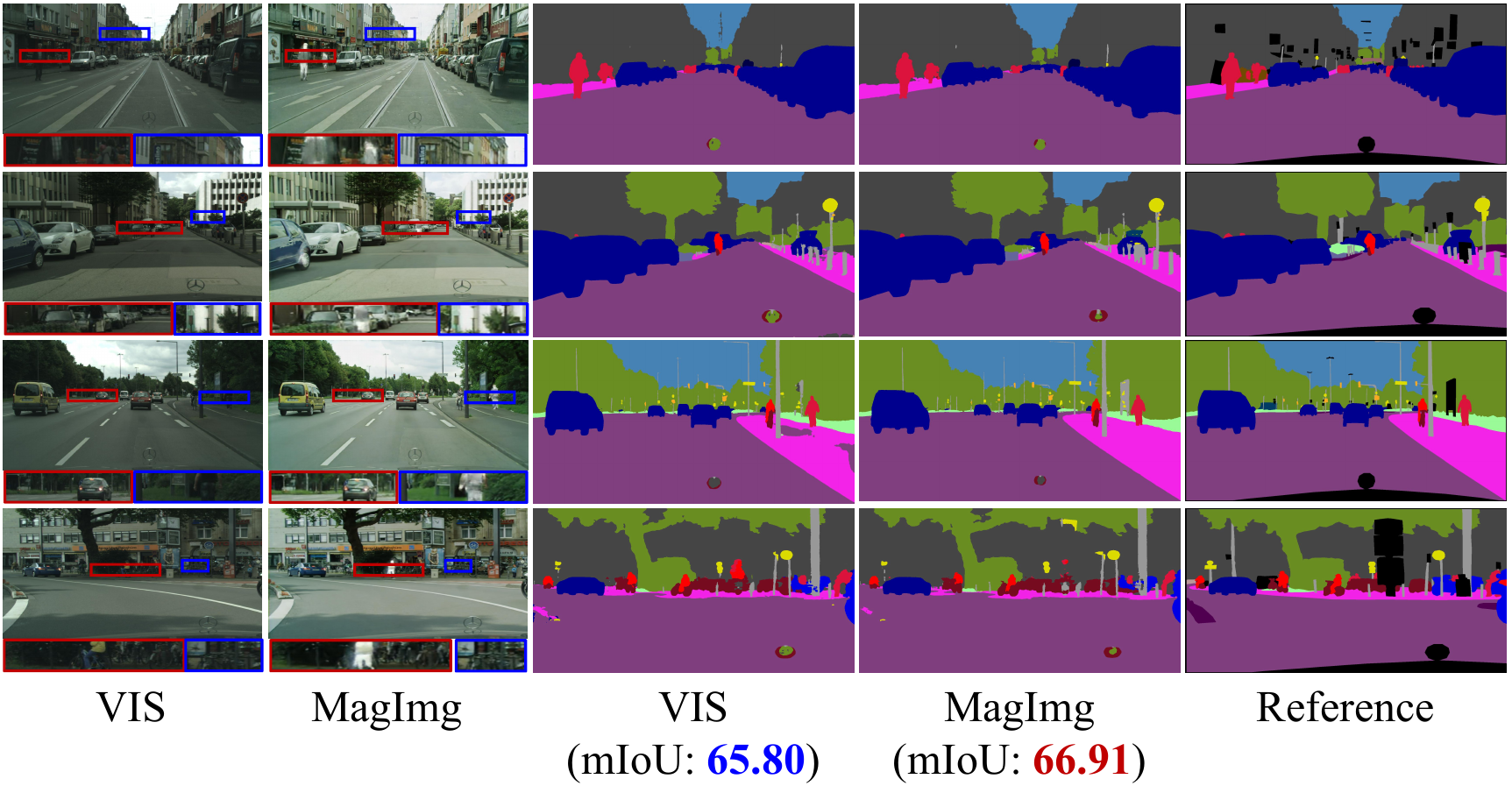}
	\vspace{-0.27in}
	\caption{Applying our MagicFuse beyond fusion scenarios.}\label{fig:9}
	\vspace{-0.10in}
\end{figure}

\subsection{Beyond Fusion Scenarios}
Previous experiments are implemented on the infrared-visible fusion datasets. Since MagicFuse requires only a single visible image, it can be applied to nearly any natural image. To demonstrate this, we evaluate it on the Cityscapes dataset~\cite{cordts2016cityscapes}. Specifically, $490$ visible images are enhanced to generate MagImgs for retraining SegFormer, and evaluation is conducted on $154$ MagImgs. As shown in Fig.~\ref{fig:9}, MagicFuse improves scene visibility and boosts semantic segmentation mIoU from $65.80$ to $66.91$, demonstrating its robust capability in enhancing scene representations.

\begin{table}[t]
	\caption{Quantitative results of visual ablation studies.} \label{tab:6}
	\vspace{-0.1in}
	\centering
	\scriptsize 
	\setlength{\tabcolsep}{10pt}
	\renewcommand{\arraystretch}{1}
	\begin{tabular}{@{}lccccc}
		\toprule
		\textbf{Variants} & \textbf{EN}$\uparrow$ & \textbf{SSIM}$\uparrow$ & \textbf{MI}$\uparrow$ & \textbf{Qabf}$\uparrow$ & \textbf{PSNR}$\uparrow$ \\ \midrule
		\textbf{Model I} & 7.28 & 0.43 & 4.09 & \cellcolor[HTML]{FFE6E6}\textbf{0.53} & 62.14 \\
		\textbf{Model II} & \cellcolor[HTML]{E6E6FF}{\ul{7.33}} & 0.43 & \cellcolor[HTML]{E6E6FF}{\ul 4.10} &  0.50 & \cellcolor[HTML]{E6E6FF}{\ul 62.16} \\
		\textbf{Model III} & 7.17 & 0.41 & 3.24 & 0.49 & 61.49 \\
		\textbf{Model IV} &\cellcolor[HTML]{FFE6E6}{\textbf{ 7.39}}  &\cellcolor[HTML]{E6E6FF}{\ul 0.44}  &4.08  &0.50  &62.11  \\
		\textbf{Full Model} & 7.29 & \cellcolor[HTML]{FFE6E6}\textbf{0.45} & \cellcolor[HTML]{FFE6E6}\textbf{4.13} & \cellcolor[HTML]{E6E6FF}{\ul 0.50} & \cellcolor[HTML]{FFE6E6}63.49 \\ \bottomrule
	\end{tabular}
	\vspace{-0.05in}
\end{table}

\begin{table}[t]
	\caption{Quantitative results of semantic ablation studies.} \label{tab:7}
	\vspace{-0.1in}
	\centering
	\scriptsize 
	\setlength{\tabcolsep}{1.6pt}
	\renewcommand{\arraystretch}{0.95}
	\begin{tabular}{@{}lcccccccc|c}
		\toprule
		\textbf{Variants} & \textbf{Car} & \textbf{Person} & \textbf{Bike} & \textbf{Curve} & \textbf{Car Stop} & \textbf{Guar.} & \textbf{Cone} & \textbf{Bump} & \textbf{mIoU}$\uparrow$ \\ \midrule
		\textbf{Model I} & 83.65 & 51.45 & 70.5 & 34.87 & 55.09 & 48.95 & 62.62 & 44.94 & 61.05 \\
		\textbf{Model II} & 83.54 & 51.31 & 70.55 & 34.87 & 54.82 & 47.78 & 62.7 & 44.45 & 60.83 \\
		\textbf{Model III} & 79.92 & 59.81 & 68.45 & 32.96 & 50.03 & 43.01 & 61.65 & 44.59 & 59.74 \\
		\textbf{Model IV}& 85.99 & 69.06 & 70.79 & 25.49 & 42.98 & 7.15 & 56.61 & 58.93 & 57.18 \\
		\rowcolor[HTML]{FFE6E6} 
		\textbf{Full Model} & 83.80 & 61.24 & 70.62 & 35.28 & 54.09 & 49.42 & 62.69 & 45.09 & \textbf{62.19} \\ \bottomrule
	\end{tabular}
	\vspace{-0.1in}
\end{table}

\section{Conclusion}
This paper proposes MagicFuse, a single-image fusion framework designed for robust scene representation under harsh conditions using only a degraded visible image. It uses an IKR branch to recover obscured visible information and a CKG branch to learn infrared-aligned thermal patterns. Then, an MKF branch is developed to integrate probabilistic noise from both branches into a unified diffusion stream, combining multi-domain knowledge. Guided by visual and semantic constraints, MagicFuse produces representations suitable for both human perception and semantic tasks. Experiments show that our method, relying solely on a single degraded visible image, achieves performance on par with or surpassing SOTA multi-modal fusion methods.

\section{Acknowledgments}
This work was supported by NSFC (62506268, 62276192), the Basic Research Program of Jiangsu (BK20250454), and CPSF (GZB20250066, 2025M781505).  
  
{
	\newpage
    \small
    \bibliographystyle{ieeenat_fullname}
    \bibliography{main}
}


\end{document}